\documentclass[pdflatex,sn-mathphys]{sn-jnl}
\usepackage{graphicx}%
\usepackage{multirow}%
\usepackage{amsmath,amssymb,amsfonts}%
\usepackage{amsthm}%
\usepackage{mathrsfs}%
\usepackage[title]{appendix}%
\usepackage{xcolor}%
\usepackage{textcomp}%
\usepackage{manyfoot}%
\usepackage{booktabs}%
\usepackage{algorithm}%
\usepackage{algorithmicx}%
\usepackage{algpseudocode}%
\usepackage{listings}%
\usepackage{changepage}
\usepackage{lmodern}
\usepackage{fix-cm}
\usepackage{lmodern} 
\usepackage[T1]{fontenc}
\usepackage{lmodern}  % Load the Latin Modern font family
\renewcommand{\bfseries}{\fontseries{b}\selectfont}  % Ensure bold series is defined
\usepackage{booktabs}
\jyear{2021}%
\theoremstyle{thmstyleone}%
\theoremstyle{thmstyletwo}%
\theoremstyle{thmstylethree}%
\begin{document}
%\begin{frontmatter}
\title[Research Article]{An Advanced Deep Learning Based Three-Stream Hybrid Model for Dynamic Hand Gesture Recognition}


\author*[1]{\fnm{Md Abdur} \sur{Rahim}}\email{rahim@pust.ac.bd}

\author*[2]{\fnm{Abu Saleh Musa} \sur{Miah}}\email{abusalehcse.ru@gmail.com}

\author[1]{\fnm{Hemel Sharker} \sur{Akash}}\email{hemelakash472@gmail.com}

\author*[2]{\fnm{Jungpil} \sur{Shin}}\email{jpshin@u-aizu.ac.jp}
\author[3]{\fnm{Md. Imran} \sur{Hossain}}\email{imran05ice@pust.ac.bd}
\author[4]{\fnm{Md. Najmul} \sur{Hossain}}\email{najmul\_eece@pust.ac.bd}
\affil[1]{\orgdiv{Department of Computer Science and Engineering}, \orgname{Pabna University of Science and Technology}, \orgaddress{\street{Rajapur}, \city{Pabna}, \postcode{6600}, \country{Bangladesh}}}
\affil[2]{\orgdiv{School of Computer Science and Engineering}, \orgname{The University of Aizu}, \orgaddress{\street{Aizuwakamatsu}, \city{Fukushima}, \postcode{9658580}, \country{Japan}}}
\affil[3]{\orgdiv{Department of Information and Communication Engineering}, \orgname{Pabna University of Science and Technology}, \orgaddress{\street{Rajapur}, \city{Pabna}, \postcode{6600}, \country{Bangladesh}}}
\affil[4]{\orgdiv{Department of Electrical, Electronic, and Communication Engineering}, \orgname{Pabna University of Science and Technology}, \orgaddress{\street{Rajapur}, \city{Pabna}, \postcode{6600}, \country{Bangladesh}}}
\abstract{In the modern context, hand gesture recognition has emerged as a focal point. This is due to its wide range of applications, which include comprehending sign language, factories, hands-free devices, and guiding robots. Many researchers have attempted to develop more effective techniques for recognizing these hand gestures. However, there are challenges like dataset limitations, variations in hand forms, external environments, and inconsistent lighting conditions. To address these challenges,  we proposed a novel three-stream hybrid model that combines RGB pixel and skeleton-based features to recognize hand gestures. In the procedure, we preprocessed the dataset, including augmentation, to make rotation, translation, and scaling independent systems. We employed a three-stream hybrid model to extract the multi-feature fusion using the power of the deep learning module. In the first stream, we extracted the initial feature using the pre-trained Imagenet module and then enhanced this feature by using a multi-layer of the GRU and LSTM modules. In the second stream, we extracted the initial feature with the pre-trained ReseNet module and enhanced it with the various combinations of the GRU and LSTM modules. In the third stream, we extracted the hand pose key points using the media pipe and then enhanced them using the stacked LSTM to produce the hierarchical feature. After that, we concatenated the three features to produce the final. Finally, we employed a classification module to produce the probabilistic map to generate predicted output. We mainly produced a powerful feature vector by taking advantage of the pixel-based deep learning feature and pos-estimation-based stacked deep learning feature, including a pre-trained model with a scratched deep learning model for unequalled gesture detection capabilities. The design of the proposed system is intended to be useful in challenging industrial situations and to create efficient, contactless interfaces. We conducted extensive experiments with the newly created hand-gesture dataset, and the proposed model achieved good performance accuracy.}

\keywords{Hand Gesture, Non-Touch Interface, LSTM, GRU, Hybrid Model, Artificial Intelligence, Deep Learning Model}

\maketitle
\section{Introduction}\label{sec1}
The gesture is a significant means of nonverbal communication involving body parts to convey \cite{rangaswamy2020vepad} important information. This information is used in conjunction with spoken words instead of speaking and to provide instructions in an uncontrolled environment. Gestures include a variety of movements that can be performed using different body parts, such as the hands, face, or body. In recent years, human-computer Interaction (HCI) has received significant attention, especially in gesture recognition, since the hand is the most convenient and direct way of expressing our feelings compared to the other parts of the body \cite{bib1}. Many researchers have been working on hand gesture research, including sign language recognition \cite{miah2022bensignnet_miah,miah2024sign_largescale_miah,miah2024hand_multiculture_miah,miah2024spatial_paa_paa,10360810_ksl2_miah,shin2023korean_miah,shin2023rotation_miah}, air writing \cite{bib3}, non-touch interfaces, robotic instructions, and pose estimation \cite{bib4}. A gestural flick input-based character input system was proposed in \cite{bib5}. The authors used body skeleton information to recognize hand gestures for Japanese hiragana, English, and numeric inputs. However, the distance for recording the body index is a critical challenge for improved recognition. In \cite{bib6,shin2023rotation_miah}, a sign language word was detected using the hybrid segmentation method on a hand motion. The YCbCr and SkinMask segmentation techniques were employed, as well as feature fusion of the support vector machine (SVM) for training and classification. The luminance of the lighting could affect the recognition of sign word gestures. The hand gestures were recognized using a fine-tuned CNN approach \cite{bib7}. For the performance evaluation, the author used two independent datasets and a real-time American sign language. However, hand shape and image processing overhead act as important factors. \\
Vision-based hand gesture recognition encompasses several stages, namely data acquisition, hand region segmentation, feature extraction, and gesture classification using the extracted features \cite{bib8}. Different sensor devices, including Kinect, web cameras, RGBD sensors, and leap motion, are used for data acquisition \cite{miah2024hand_multiculture_miah,miah2024effective_emg,miah2023skeleton_euvip,miah2023dynamic_mcsoc}. A method that relies on the analysis of human skin colour is used to accurately identify and locate hands in colour images, operating under the assumption that the hand region predominantly occupies the frame of the image \cite{bib9,miah2024review}. Hand segmentation can be challenging when the hand is close to the human face or body and the background colour closely resembles the human skin colour. In addition, researchers have applied handcrafted feature extraction techniques such as shape descriptors, spatial features and hand gesture recognition \cite{bib10}. These characteristics improve performance in a certain context. Recently, image processing and vision-based gesture recognition using deep learning techniques such as CNN have obtained promising results \cite{bib11}. It has required a huge number of image datasets and high memory resources to train effectively. In this study, we enlarge our dataset using data augmentation techniques.   \\
The increasing prevalence of sensors and cameras in technology is leading to a rise in non-touch devices and interfaces in our daily lives. In recent years, the field of Human-Computer Interaction (HCI) has made significant advancements in improving the effectiveness of gesture recognition applications. Hand gesture recognition is a crucial technique for advancing HCI \cite{10529244_miah_ksl0,kabir2024bangla_miah,hassan2024residual_miah_alzh,10510436_miah_anomaly}. It offers users a natural and direct means of expressing their emotions, interacting with VR/AR devices, and engaging in human-machine interaction and gaming \cite{bib12,bib13}. Numerous previous investigations have focused on the recognition of device-based gestures. Specialized hardware and dedicated devices or peripheral devices are necessary for this type of processing. However, these technologies can be inconvenient and challenging to use in real-world scenarios due to the requirement of direct contact with additional devices by the users. \\
In this paper, we collected hand gesture-based videos and separated the gesture images for training and testing purposes. To improve the image quality, we preprocess the input images and apply augmentation techniques to enlarge the dataset images. The main contributions of the works are as follows.
\begin{itemize}
    \item [i.] \textbf{Novelty}: We created a comprehensive dataset using continuous video recordings that feature ten distinct hand gestures. This dataset is pivotal, serving as the foundation for both the training and validation of our hand gesture recognition models. Then we used Recognizing the inherent challenges in hand gesture datasets, such as variations due to rotation, movement, and scale, we implemented thorough preprocessing techniques. Additionally, we augmented our dataset to ensure it's robust and can effectively train our model, preparing it for real-world variations. Then we employed deep learning based three stream feature extraction technique.  
 
    \item [ii.] \textbf{Pixel-Based Deep Learning Features }: In the first and second streams, we utilized pixel-based deep earning features enhancement. 
    In the first stream, we extracted the initial feature using the pre-trained Imagenet module and then enhanced this feature by using a multi-layer of the GRU and LSTM modules. In the second stream, we extracted the initial feature with the pre-trained ReseNet module and enhanced it with the various combinations of the GRU and LSTM modules. 
    
    \item [iii] \textbf{Joint Skeleton Based Stacked-LSTM Features}
    In the third stream, we extracted the hand pose key points using the media pipe and then enhanced them using the stacked LSTM to produce the hierarchical feature. After that, we concatenated the three features to produce the final. 
    
    \item [iv.] \textbf{Classification and Enhancement}:
    Finally, we employed a classification module to produce the probabilistic map to generate predicted output. We mainly produced a powerful feature vector by taking advantage of the pixel-based deep learning feature and pos-estimation-based stacked deep learning feature, including a pre-trained model with a scratched deep learning model for unequalled gesture detection capabilities.  In rigorous evaluations using our custom dataset, our model achieved a noteworthy average accuracy of 98.35\%. This achievement places our method in a competitive position against state-of-the-art solutions and indicates its superiority over individual strategies.
\end{itemize}

This paper is organized as follows. Section 2 describes relevant works of recent techniques by various scientists. The proposed methodology, which includes dataset description, image processing and augmentation, feature extraction and classification, is discussed in Section 3. Section 4 describes the experimental results obtained using the proposed methods. Finally, Section 5 concludes this paper.  

\section{Related Works}\label{sec2}

This section presents a comprehensive literature review of recent strategies for vision-based hand gesture recognition. The research involves hand gesture recognition using RGB cameras and depth sensors, as well as machine learning and deep learning techniques. A feature fusion-based CNN is used to recognize the hand posture in \cite{bib14,miah2023dynamic_miah,mallik2024virtual_miah,shin2024japanese_jsl1_miah}. The authors conducted experiments using binary, grayscale, and depth data, along with various evaluation techniques. Four types of characteristics are recovered from the input hand gesture images: Hu moments, Zernike moments (ZM), Gabor filter, and contour features. The aforementioned four CNN-structured features are combined at the final fully connected layers to provide a final gesture class. In \cite{bib15}, the authors conducted a review of different methods and databases for hand gesture recognition. A quantitative and qualitative comparison of algorithms was performed on RGB and RGB-D cameras. Chen et al. \cite{bib16} tracked hands in real time and recovered hand shapes from varied backgrounds using Fourier descriptors and optical flow-based motion analysis. This reduces the possibility of gesturing with the model varying the gesture from a reference pattern. Lee et al. \cite{bib17} suggested a unique technique for static gesture recognition in challenging circumstances based on wristband-based contour data. To precisely segment the hand region, a set of black wristbands is placed on both hands. However, wearing wristbands all the time may be inconvenient and hazardous. In the context of static hand gesture recognition, a multi-objective evolutionary algorithm is employed to optimize a composite of attributes and dimensions effectively. A recognition accuracy of up to 97.63\% was achieved on a dataset consisting of 36 gesture poses from the MU dataset \cite{bib18}. \\
However, the hand region segmentation problem can be effectively addressed through the use of an RGBD sensor. The subsequent literature pertains to the field of hand gesture recognition using RGB-D sensor technology. In \cite{bib19,10399807_miah_blind_writing,egawa2023dynamic_miah} the authors proposed a deep learning technique used to recognize gestures and postures to assist machines in various fields. In \cite{miah2023dynamic_miah}, the researchers employed a multi-branch attention-based graph and deep learning model to identify hand gestures accurately. This was achieved by extracting a comprehensive range of skeleton-based features. The authors achieved 97.01\% accuracy using the SHREC’17 dataset. Ma et al. \cite{bib121} proposed a modified memory-augmented neural network known with Long Short-Term Memory (LSTM) architecture as a means to enhance the efficiency of the system. The researchers achieved accuracy rates of 82.29\% and 82.03\% for the DHGD dataset and 79.17\% for the MSRA dataset. A soft-voting model-based Bengali Sign gesture recognition system was proposed in \cite{bib122}. The authors proposed a Bengali Sign dataset and achieved the average accuracy using the CNN model. After analyzing different hand gestures using preprocessing and feature extraction processes, it is possible to achieve the goal using deep learning technology.

\section{Proposed Methodology}\label{sec3}
The primary goal of this research is to obtain a high level of features to increase the performance accuracy in hand gesture recognition with low effort and expense by employing vision-based image and video-based hand gesture recognition. Figure 1 demonstrates the workflow architecture of the proposed method. The proposed methodology involved in our research on hand gesture recognition begins with dataset preprocessing and augmentation to ensure uniformity and diversity in training samples. We then develop a novel three-stream hybrid model that combines RGB pixel and skeleton-based features for enhanced recognition. In the feature extraction stage, we utilize pre-trained Imagenet and ResNet modules in the first two streams, enhancing features with GRU and LSTM layers. The third stream extracts hand pose key points using Mediapipe and enhances them with stacked LSTM for hierarchical feature generation. Figure 6 demonstrates the details of the three streams. These features are then fused to create a final representation. A classification module is employed to generate a probabilistic map for predicting outputs. The proposed methodology is developed based on both still images and video sequences, all while ensuring reduced effort and cost. Our methodology commences with a thorough analysis of hand gesture videos to grasp the full spectrum of gestures we are dealing with. Following this, individual gesture frames are meticulously segmented to spotlight the gesture itself, subtracting it from superfluous background elements. With segmentation complete, the images are subjected to a rigorous preprocessing regimen to foster uniformity. Concurrently, to combat real-world variability and enrich our dataset, we introduce controlled changes through augmentation. The pivotal stage of feature extraction then follows. Here, by leveraging powerful models such as ImageNet, ResNet50V2, and Mediapipe, we draw out salient gesture-specific attributes from each image or frame, emphasizing hand pose, hand position, and finger position dynamics. Armed with these distilled features, our next move is to deploy a sophisticated hybrid model, which elegantly combines the strengths of transfer learning with LSTM and GRU to decipher and categorize specific hand gestures. 

\begin{figure}[ht]
\centering
\includegraphics[scale=.4]{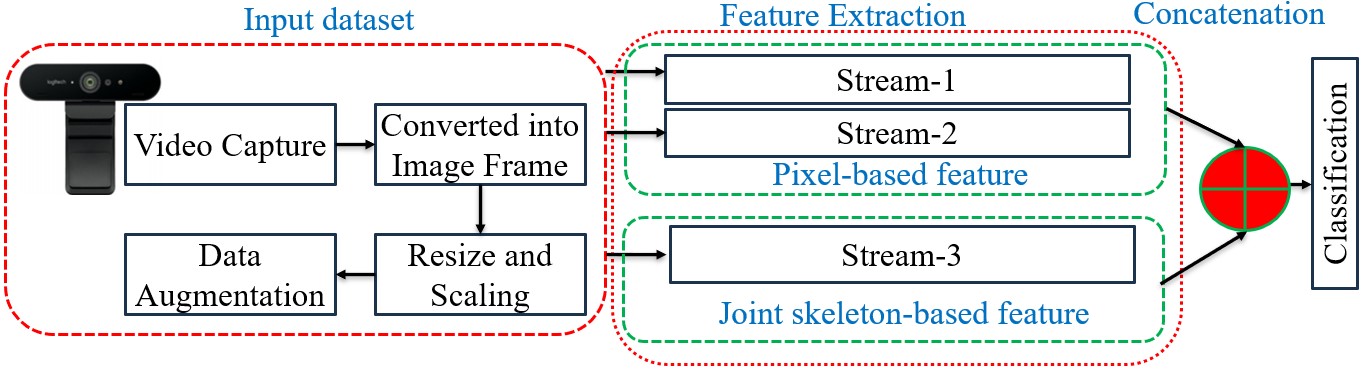}
\caption{Basic working flow of the proposed hand gesture recognition system.}\label{fig1}
\end{figure}

\subsection{Dataset Description} \label{subsec3}
In this work, we proposed a hand gesture dataset for evaluating the gesture recognition performance. It contains a total of ten different dynamic hand gestures, for example, left, right, up, down, hi, bye, open, close, thumbs up, and thumbs down. We collected this data using the webcam (Logitech BRIO Ultra HD Pro 4K). Seventeen (17) individuals were asked to perform dynamic hand gestures to collect the dataset. Each person completed each activity five times, which were video. It has been recorded in various locations and lighting circumstances. The video's resolution was 1080x1920, with a frame rate of 30 FPS. We collected a maximum of 1500 images for ten (10) tasks per person and 150 images for each gesture. In this case, a total of 25500 original dataset images were collected. Therefore, we preprocessed the input images to evaluate performance and used the augmentation technique to enlarge the dataset images. Figure 2 shows an example of the input images of the proposed dataset. Table 1 represents the descriptions of the hand gesture actions. The is available online at this URL: \url{https://drive.google.com/drive/folders/16FKgQ2Y-PuMu-hNp-sjww0PSJppRR204?usp=sharing}

\begin{figure}[ht]
\centering
\includegraphics[scale=.98]{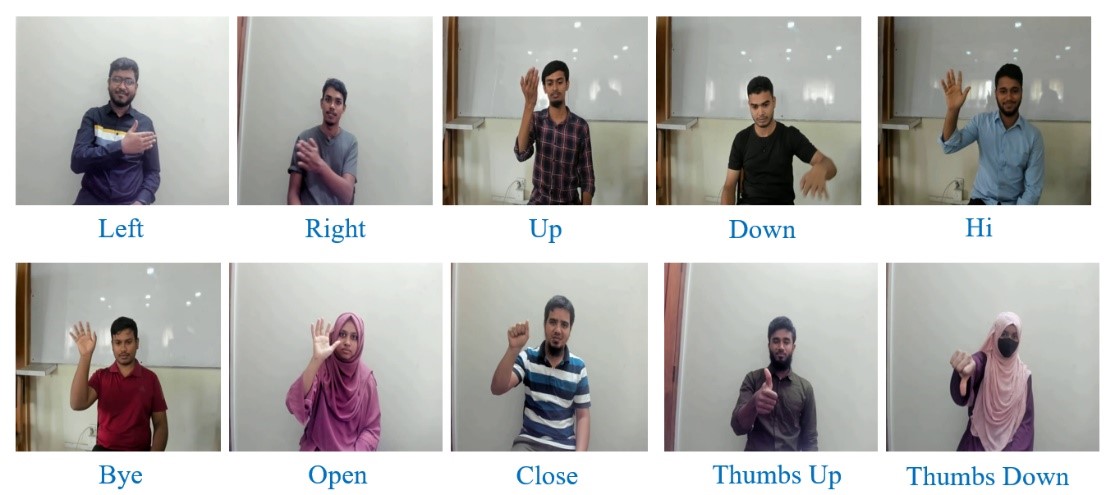}
\caption{Example of the input images.}\label{fig2}
\end{figure}

\begin{table}[ht]
\caption{Description of hand gesture actions.}\label{tab1}%
\begin{tabular}{@{}ll@{}}
\toprule
Hand Gesture Actions 	& Description \\
\midrule
Left	& Move right hand from right to left \\ 
Right	& Move left hand from left to right \\
Up	& Move right or left hand from down to up \\
Down	& Move right or left hand from up to down \\
Hi	& Wave hand near side of head \\
Bye	& Move the open hand to the side of the head to the left and right \\
Open	& Hold hands to show all fingers \\
Close	& All fingers are fists \\
Thumbs up	& Hold the hand with the thumb pointing up \\
Thumbs down	& Hold the hand with the thumb pointing down \\
\botrule
\end{tabular}
\end{table}

\subsection{Preprocessing} \label{subsec3.2}
Data preprocessing was conducted to transform the datasets into a format that is suitable for subsequent analysis, ensuring usability and comprehensibility. The dataset was exposed to a resizing operation, resulting in a uniform resolution of 128x128 pixels. Additionally, the image values were scaled from the original range of 0-255 to a normalized range of 0-1. 

\subsection{Data Augmentation} \label{subsec3.3}
Augmentation techniques were employed to expand the dataset \cite{bib123} artificially. The proposed approach encompasses the implementation of minor modifications to the existing dataset, leveraging deep learning techniques to generate novel data instances. We applied three augmentation techniques in this process: brightness, horizontal flip, and rotation (10 degrees) respectively. A horizontal flip is a geometric transformation that reflects an image across the vertical axis, resulting in a mirrored representation of the original image. This process effectively swaps the positions of pixels along the horizontal axis, thereby creating a visual effect of the image being flipped from left to right or across the screen as the dataset image is split into 80\% and 20\% for training and testing purposes. Therefore, we further augment only the training images for rotation and brightness. The brightness effect was employed to alter the intensity of the input image being acquired. In addition, the dataset undergoes a random clockwise rotation of 10 degrees. After augmentation, the dataset images for training are 132600 and a total of 10200 images for testing.

\subsection{Feature Extraction and Classification} \label{subsec3.4}
The evaluation process for our proposed hybrid model requires extracting features from the input dataset. In this context, we utilized three distinct feature extraction techniques that is considered as three streams, namely ImageNet, ResNet50V2, and Mediapipe. The hybrid models we propose are acknowledged as potent solutions for image classification challenges, as referenced in \cite{bib124}. An integral part of our methodology is the pre-training of these models to address new tasks adeptly. Within this research framework, ImageNet and ResNet50V2 are specifically employed to scrutinize feature extraction capabilities on a multi-class hand gesture dataset. Crucially, to attain the desired results, we harnessed the time-distributed layer available in both the ImageNet and ResNet50V2 architectures. The rationale behind this is underscored by the layer's ability to extract features from individual image frames effectively. Furthermore, to align with our specific classification task, we replaced the terminal fully connected (FC) dense layer in each network with a new one consisting of 'n' nodes. Here, 'n' equates to the total number of gesture classes represented in our dataset.

\subsubsection{ImageNet} \label{subsec3.4.1}
ImageNet is a large-scale image library that is commonly used to train and test computer vision models \cite{bib125}. ImageNet-trained models performed well at a variety of computer vision tasks, including image classification, object identification, and image segmentation. ImageNet has evolved into a typical benchmark for assessing the performance of computer vision algorithms. Many advanced algorithms are frequently compared based on their performance on the ImageNet validation dataset. In addition to growing in size and variety, the dataset now contains more images and categories in its most recent variants. ImageNet's model is depicted in Figure 3.  

\begin{figure}[ht]
\centering
\includegraphics[scale=.75]{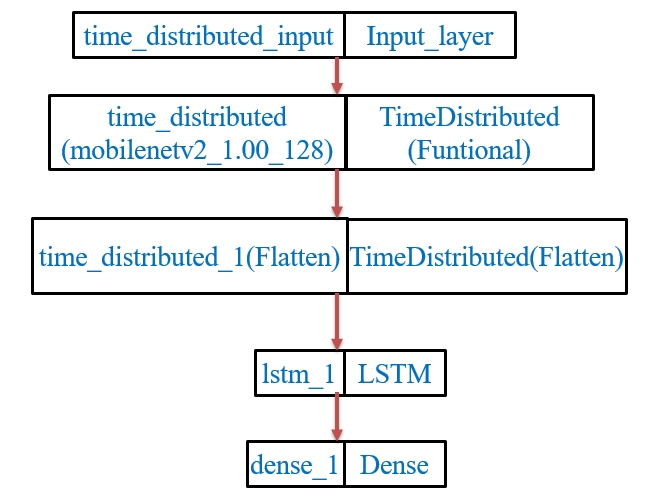}
\caption{The architecture of the ImageNet.}\label{fig3}
\end{figure}

\subsubsection{Resnet50v2} \label{subsec3.4.2}
ResNet-50 is a widely used and highly effective pertained deep-learning model specifically designed for image classification tasks \cite{bib126}. It is based on CNNs, which are a type of deep neural network frequently used for analyzing visual images. It has 50 layers with 48 convolutional layers, 1 MaxPool and one average pool layer. This study used two time-distributed ResNet50v2 models. Figure 4 shows the ResNet50 model architecture. It has two input types, but no type length. However, the first input image length is 30, and the other frame is 5. The second input frame considers the length of 0, 7, 15, 22, 29 positional frames that help the entire model to classify the hand gesture and improve the accuracy. 

\begin{figure}[ht]
\centering
\includegraphics[scale=.55]{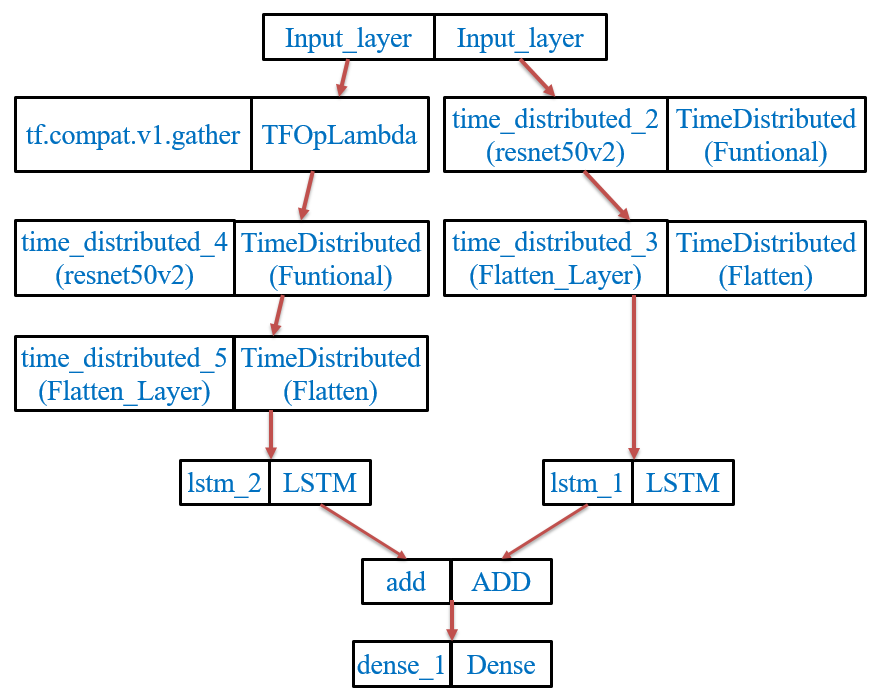}
\caption{ResNet50 model architecture.}\label{fig4}
\end{figure}

\subsubsection{Landmark Information with Media pipe} \label{subsec3.4.3}
By analyzing input data, Mediapipe applies a machine-learning algorithm to estimate hand joints. The initial job of the model is to recognize the existence of a hand within the camera image and then calculate the 3D coordinates of the hand joints \cite{bib127}. This procedure entails complex mathematical computations based on input data such as the hand's position and orientation, as well as variables such as lighting conditions and camera angle. Following these calculations, the model generates a set of coordinates corresponding to the estimated positions of each hand joint in 3D space. For each image, we estimated 258 positions. For pose estimations, there are 33 points. Each point has four characteristics: X, Y, Z, and Visibility. As a result, the total number of hand pose estimation points is 132.

Furthermore, both left and right-hand landmarks have 21 points. The landmarks have three properties such as X, Y, and Z. As a result, the total number of landmark positions for both the left and right-hand landmarks is 63. Figure 5 shows an example of landmark position detection of an input image. 

\begin{figure}[ht]
\centering
\includegraphics{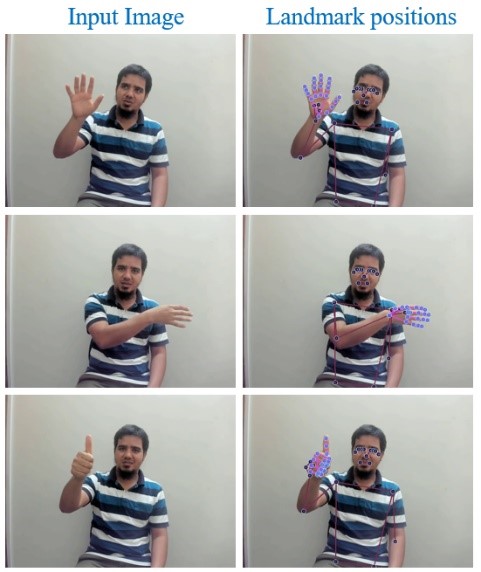}
\caption{Example of the landmark information of an input image.}\label{fig5}
\end{figure}

\subsubsection{Long Short-Term Memory (LSTM)} \label{subsec3.4.4}
Long Short-Term Memory (LSTM) is a form of recurrent neural network (RNN) architecture that is used to capture long-range dependencies in sequential data. It is widely used in natural language processing, speech recognition, time series analysis, and other applications \cite{miah2023dynamic_miah}. Three LSTM gates are involved in the information flow into and out of the memory cell. The Forget Gate is the first gate, and it decides whether or not to discard data from the previous cell state. It generates a value between 0 and 1 for each cell state element after accepting the current input and the prior cell state as inputs. For example, a value of 0 means "forget" while a value of 1 means "retain." Second, the input gate determines what new information to include in the cell state. \\
Additionally, it takes the current input and the previous cell state as inputs, and it produces a candidate cell state with values that can be appended to the cell state. Third, the output gate decides what information should be delivered as the LSTM's output. The current input and the updated cell state are used as inputs to produce the output. The ability of LSTMs to recall is critical in hand gesture detection. Because the preceding frames establish the context, the understanding of the frames that follow may be influenced. \\
In most cases, an LSTM layer is added after convolutional layers to extract features or an embedding layer to manage structured data. Stacking numerous LSTM layers may be advantageous in comprehending sophisticated patterns and improving the accuracy in recognising movements. Following the data's move through the LSTM, the output layer is tasked with predicting the precise sign or gesture. One limitation of this system is the inconsistency of the length of the sign gestures. Techniques such as padding are required to manage sequences of various lengths. For real-time applications, it is necessary to find a balance between the model's complexity and its operating speed. In essence, given their ability to handle sequential data, LSTMs stand out as a promising method for efficient sign language detection, especially when combined with other neural network components.

\subsubsection{Gated Recurrent Units (GRU)} \label{subsec3.4.5}
Gated Recurrent Units (GRUs) are a type of recurrent neural network (RNN) architecture that was designed to resolve some of the problems with traditional RNNs, especially the vanishing gradient problem, which makes it difficult for RNNs to figure out how sequences are related over a long time \cite{bib129}. GRUs were designed to be efficient at accomplishing computations, making the gating mechanism simpler while keeping the ability to capture and remember important information from previous time steps. The GRU basically has two gates: the Reset Gate, which decides what information from the past to throw away, and the Update Gate, which decides how much information from the previous state to use. This structure makes it easier to recognize patterns over time in data. The following Equations (1-4) regulate the behaviour of a GRU unit.

\begin{equation}
Z_t=sigmoid(W_z\times [h_{t-1},x_t])
\end{equation}

\begin{equation}
r_t=sigmoid(W_t\times [h_{t-1},x_t])
\end{equation}

\begin{equation}
h_t=tanh(W_h \times [h_{t-1},x_t])
\end{equation}

\begin{equation}
h_t=(1-z_t) \times h_{t-1}+Z_t \times h_t])
\end{equation}

Where $x_t$ is the input at time step $t$ and $W_z$, $W_r$, and $W_h$ are the weight matrices that are gained during learning.

\subsubsection{Features Explanation of the Proposed Hybrid Model} \label{subsec3.4.6}
Three streams are used to compose the proposed hybrid mode. In each stream, we used a hybrid module that combines the strengths of both LSTM and GRU, allowing the model to capture long-term dependencies with LSTM while benefiting from the computational efficiency of GRU. It adjusts the number of LSTM and GRU units, as well as other hyperparameters, based on your specific dataset and requirements. Figure 6 depicts the architecture of the proposed hybrid model. \\

\textbf{First Stream}: The foundational architecture of this stream is anchored by the ImageNet module. ImageNet, a vast database of labelled images, serves as a cornerstone, aiding in the extraction of generalized and discriminative features that enhance the robustness of the model. Following this, the features are bifurcated into two branches. The first branch merges the capabilities of LSTM and GRU, which efficiently handles both short-term and long-term dependencies, complemented by a Dense Layer that introduces depth, allowing the network to discern and learn intricate patterns. Simultaneously, the second branch employs a Gather operation that effectively aggregates features, and three LSTM modules ensure the comprehensive capture of sequential data over extended intervals. \\
\textbf{Second Stream}: Initiating this stream is the ResNet50V2 architecture. ResNet50V2, renowned for its residual connections, deftly addresses the vanishing gradient problem, facilitating the creation of deeper networks and leading to enhanced feature extraction without informational loss. These features are then split into two branches. The first integrates two GRU modules known for computational efficiency and mitigating the vanishing gradient dilemma, a Dropout layer for model generalization, and a Dense Layer that amplifies model intricacy. In juxtaposition, the second branch, mirroring the strategy of the first stream, ensures consistency in processing and fortifies the robust sequential data capture across both streams. \\

\textbf{Third Stream}: A distinctive feature of this stream is the deployment of Mediapipe for hand pose estimation. Mediapipe, with its prowess in real-time performance, guarantees the meticulous capture of intricate hand gestures with remarkable accuracy. To process the derived joint skeleton points, the stream utilizes a sequence comprising three LSTM modules and two GRU modules. This blend not only encapsulates the sequential data with all its nuances but also ensures an exhaustive representation of the gestures in question. We extracted features using three streams based on the ImageNet, ResNet50v2, and the Mediapipe. In the context of the first stream with ImageNet and the second stream with Resnet50v2, it is observed that every image is associated with 2048 distinct features. Consequently, when considering each activity and individual, the total number of features relates to 307200. Additionally, the third stream with hand pose based on Mediapipe was employed to assess the human pose, hand position, and finger position. We obtained 258 positions for each image, and 38700 features were extracted in total for each action. \\

\textbf{Finally}, we concatenated the three streams' features, aiming to produce final features that fed into the dense classification layer. By synthesizing the strengths of each stream, the hybrid model emerges as a robust, efficient, and intricate mechanism for feature extraction. The collective advantages harnessed from each module prepare the model to handle complex patterns inherent in the data adeptly. The architecture of this avant-garde hybrid model is elucidated in Figure 6.

\begin{figure}[ht]
\centering
\includegraphics[scale=.95]{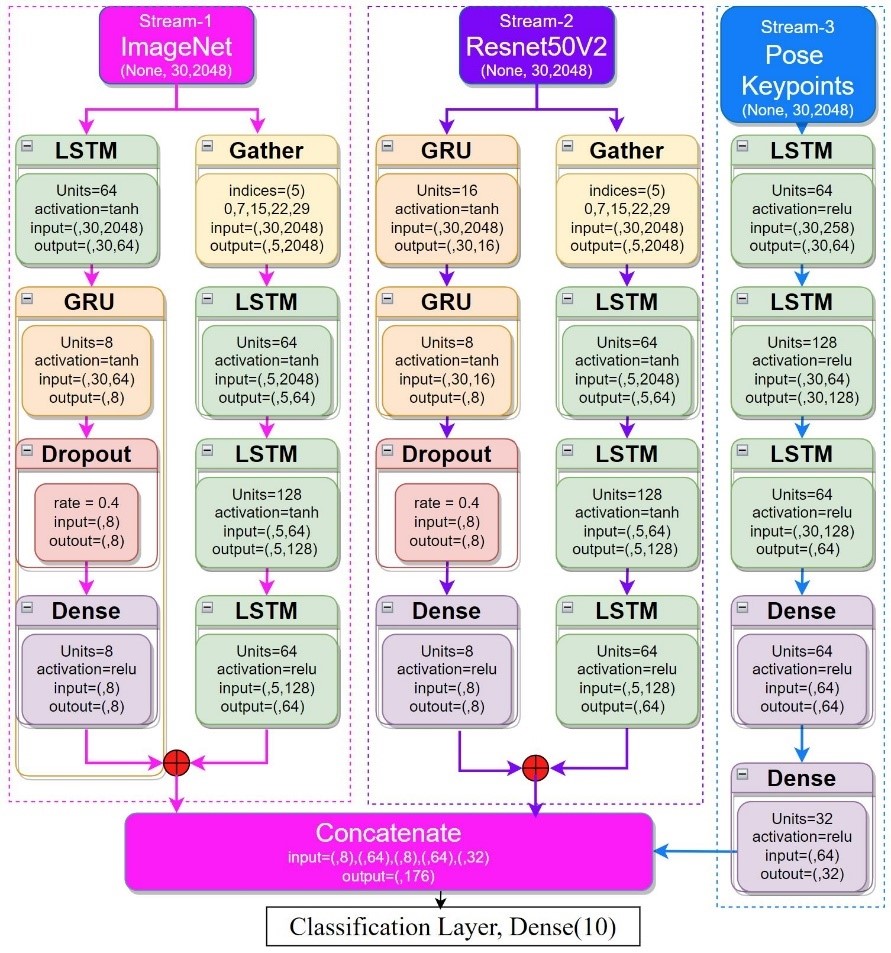}
\caption{Proposed hybrid model architecture.}\label{fig6}
\end{figure}

\section{Results and Discussion}
Experiments are conducted on a self-constructed dataset and a benchmark dataset to validate the proposed approach. To create a larger dataset, we preprocess and augment the dataset images. Consequently, three pre-trained models, namely ImageNet, ResNet50v2, and Mediapipe, were employed for feature extraction. Finally, hybrid methodologies were employed in the classification of dynamic hand gestures.
\subsection{Dataset Splitting } 
In all experiments, the datasets were partitioned into 80\% for training purposes and 20\% for testing purposes. Additionally, in order to ensure the reliability of the results, the training dataset is partitioned into a 75\% portion for training purposes and a 25\% portion for validation purposes. The dataset will now be divided into three sets: training, validation, and testing, with proportions of 60\%, 20\%, and 20\%, respectively. The training samples are used to train the model, while the validation set is employed to assess the performance of each model for hyper-parameter tuning or to determine the optimal model among various alternatives. The remaining sets designated for testing are used to assess the efficacy of our proposed model.

\subsection{Data Preprocessing, augmentation, and environment settings} 
Data preprocessing and augmentation are essential for training deep-learning models. These techniques increase the model's performance, generalization, and resilience by resolving data difficulties and limitations. In this paper, we preprocessed the input image for better performance of the proposed architecture. Therefore, augmentation techniques such as brightness, horizontal flip, and rotation (10 degrees) are applied to increase the dataset images for the proposed hybrid model. Figure 7 shows an example of the augmentation procedure using dataset images.

\begin{figure}[ht]
\centering
\includegraphics[scale=.65]{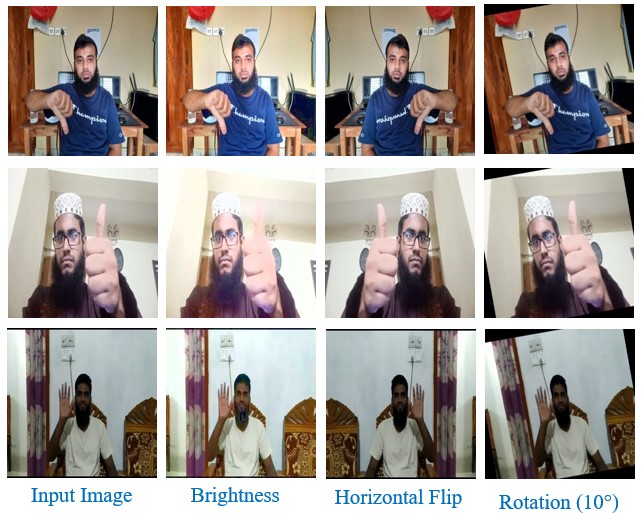}
\caption{The augmentation process of input images.\label{fig7}}
\end{figure}

The experiments were conducted in the laboratory environment, using a GeForce GTX 1660 $T_i$ Graphical Processing Unit (CPU), Central Processing Unit (CPU), and 16 GB RAM. We evaluated our data set for the 10-class classification task.

\subsection{Results and Discussion} 
In this section, we present an analysis of the outcomes derived from experiments conducted using the proposed dataset. 

\subsubsection{Results and Discussion} 
The evaluation of our proposed work's performance is conducted using accuracy (\%) as well as several evaluation metrics, including precision (PreR), recall (ReC), and F-score (FoS) values derived from the confusion matrix. The values are computed using the indices as depicted in Table 2. 

\begin{table}[ht]
\caption{Evaluation metrics of the processed system.}\label{tab2}%
\begin{tabular}{@{}ll@{}}
\toprule
Types	& Equations	\\
\midrule
PreR	& $PreR=\frac{TrP}{TrP+FaP}$ \\ 
ReC& $ReC=\frac{TrP}{TrP+FaN}$ \\ 
FoS & $FoS=\frac{2\times PreR \times ReC}{PreR+ReC}$ \\ 

Accuracy & $Accuracy=\frac{No. of correctly classified gestures}{The total number of gesture samples in the testing sets}$ \\ 

\botrule
\end{tabular}
\footnotetext{TrP=True Positive, FaP=False Positive, FaN=False Negative}
\end{table}

We extracted features using three streams based on the ImageNet, ResNet50v2, and the Mediapipe. In the context of the first stream with ImageNet and the second stream with Resnet50v2, it is observed that every image is associated with 2048 distinct features. Consequently, when considering each activity and individual, the total number of features relates to 307200. Additionally, a third stream with hand pose based on Mediapipe was employed to assess the human pose, hand position, and finger position. We obtained 258 positions for each image, and 38700 features were extracted in total for each action. Consequently, the features are employed in the training and testing phases to assess performance. 

\subsubsection{Performance Accuracy with the Newly Created Dataset} 
Figure 8 illustrates the training loss and accuracy of our proposed hybrid model. Meanwhile, Table 3 provides a detailed view of the precision, recall, f1 score and classification accuracy achieved by this model. The gestures 'Left', 'Right', 'Up', 'Down', 'Bye', and 'Thumbs Down' showcase superior performance metrics, with 'Left' achieving a precision of 97.78\% and an impeccable accuracy and recall of 100\%. Similarly, 'Right', 'Up', 'Down', and 'Thumbs Down' achieved a perfect score across all metrics, affirming the model's adeptness at recognizing these gestures. However, a closer examination reveals that some gestures, while still performing admirably, present minor challenges. For instance, the 'Open' gesture has an accuracy of 91.49\%, which is the lowest in the table. This is corroborated by its precision and recall scores of 97.73\% and 91.49\%, respectively. 'Hi' and 'Close' gestures, on the other hand, achieved comparable scores, with 'Hi' attaining an accuracy of 95.24\% and 'Close' closely following with 96.77\%. A notable aspect is the F-1 score, which provides a harmonic mean of precision and recall. The 'Thumbs Up' gesture, for instance, boasts an F-1 score of 98.46\%, indicating a well-balanced precision and recall. On aggregate, across all gestures, the model demonstrates a commendable average performance with a precision of 98.27\%, a recall of 98.35\%, an F-1 score of 98.29\%, and an overall accuracy of 98.35\%. These figures highlight the model's consistent and reliable performance in hand gesture recognition. \\
On average, the model boasts an impressive accuracy rate of 98.35\%. Notably, it achieves a perfect accuracy score of 100\% for the gestures 'left', 'right', 'up', 'down', 'bye', 'thumbs up', and 'thumbs down'. On the other hand, the 'open' gesture yielded the lowest accuracy at 91.49\%. Misclassifications were observed, especially between the 'left', 'right', and 'bye' gestures with the 'open' gesture. Moreover, gestures like 'Hi' and 'Close' also yield commendable accuracies of 95.24\% and 96.77\%, respectively. Even with gestures that posed recognition challenges, such as 'Open', the model still showcased impressive accuracy rates, underscoring its robustness and adaptability. To address these misclassification challenges, we refined our approach by focusing on specific feature models that evaluated a subset of thirty potential selection frames. Specifically, we prioritized frame numbers 0, 7, 15, 22, and 29. Our model's design emphasizes these five frames to enhance its accuracy in discerning gestures. The resulting confusion matrix, showcasing the performance of our hybrid model across different gestures, is depicted in Figure 9.

\begin{figure}[ht]
\centering
\includegraphics[scale=0.40]{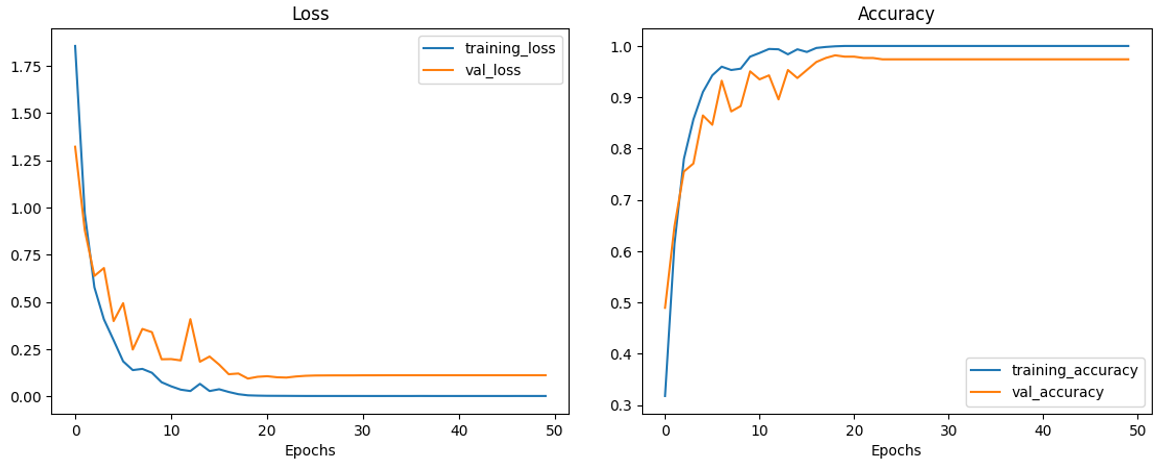}
\caption{Training loss and accuracy of the proposed hybrid model.\label{fig8}}
\end{figure}

\begin{table}[ht]
\caption{Label-wise classification results using the proposed hybrid model.}\label{tab3}%
\begin{tabular}{@{}lllll@{}}
\toprule
Gestures	& Precision (\%)	& Recall(\%)	& F-1 score (\%)	& Accuracy (\%) \\
\midrule
Left	& 97.78	& 100	& 98.88	& 100 \\ 
Right	& 100	& 100	& 100	& 100  \\ 
Up	& 100	& 100	& 100	& 100  \\ 
Down &	100	& 100	& 100	& 100  \\ 
Hi	& 97.56	& 95.24	& 96.39	& 95.24  \\ 
Bye	& 95.92	& 100	& 97.92	& 100  \\ 
Open & 	97.73	& 91.49	& 94.51	& 91.49 \\  
Close &	96.77	& 96.77	& 96.77	& 96.77  \\ 
Thumbs Up	 & 96.94	& 100	& 98.46	& 100 \\ 
Thumbs Down &	100	& 100	& 100	& 100  \\ 
Average	& 98.27	& 98.35	& 98.29	& 98.35 \\ 

\botrule
\end{tabular}
\end{table}

\begin{figure}[ht]
\centering
\includegraphics[scale=0.60]{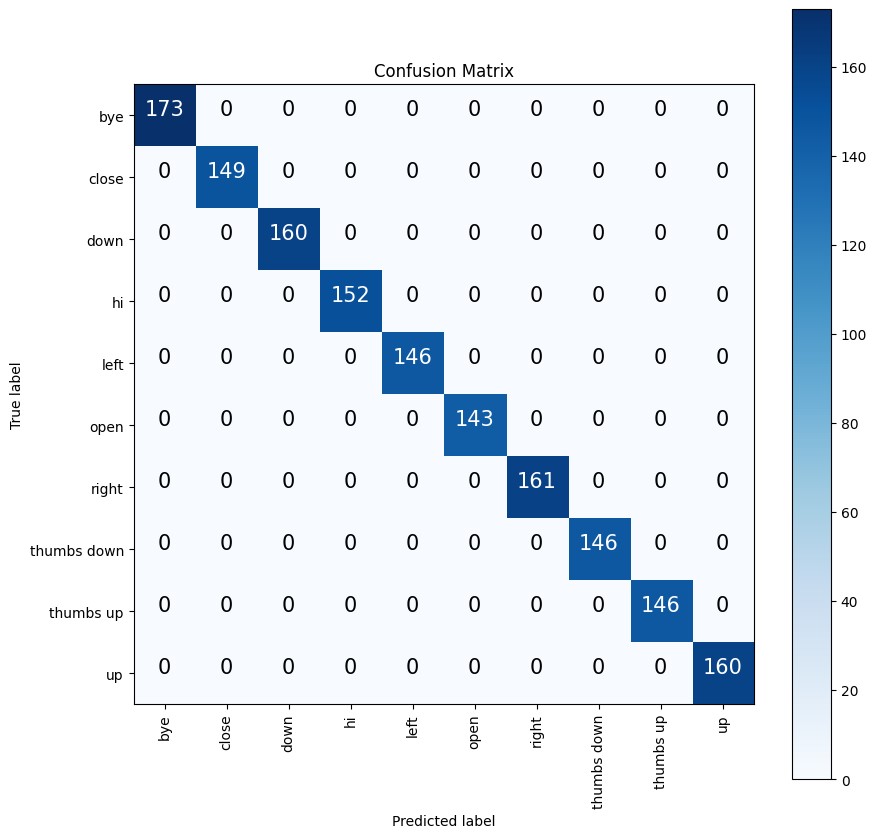}
\caption{Confusion matrix of the proposed hybrid model.\label{fig9}}
\end{figure}

\begin{table}[ht]
\caption{Classification accuracy of our proposed hybrid model on the benchmark dataset [6].}\label{tab4}%
\begin{tabular}{@{}lllll@{}}
\toprule
Gestures	& PreR (\%)	& ReC (\%)	& FoS (\%)	& Accuracy (\%) \\
\midrule
Call	& 100	&100	&100	&100 \\ 
Close	& 98.59	&1	&99.29	&98.59 \\ 
Cold	& 99.29	&98.59	&98.89	&97.04 \\ 
Correct	& 100	&100	&100	&100 \\ 
Fine	& 100	&100	&100	&100 \\ 
Help	& 98.56	&97.86	&98.21	&96.48 \\ 
Home	& 1	97.5	&98.73	&97.5 \\ 
I Love You	& 98.58	&99.29	&98.93	&97.89 \\ 
Like	& 100	&100	&100	&100 \\ 
Love	& 99.29	&99.29	&99.29	&98.59 \\ 
No	& 99.29	&99.29	&99.29	&98.58 \\ 
Ok	& 100	&100	&100	&100 \\ 
Please	& 98.58	&99.29	&98.93	&97.59 \\ 
Single	& 100	&100	&100	&100 \\ 
Sit & 	98.58	&99.29	&98.93	&97.95 \\ 
Tall	& 97.86	&98.56	&98.21	&96.48 \\ 
Wash	& 99.29	&98.58	&98.93	&97.89 \\ 
Work	& 100	&99.3	&99.65	&99.3 \\ 
Yes	& 100	&97.87	&98.9	&97.89 \\ 
You	& 99.28	& 97.87	& 98.57&	97.18 \\ 

\botrule
\end{tabular}
\end{table}

\subsubsection{Performance Accuracy with Bench Mark Hand Gesture Dataset} 
Table 4 provides an in-depth breakdown of the performance metrics for each gesture label within the benchmark hand gesture dataset. A notable achievement is the perfect accuracy score of 100\% for several gestures, including 'Call', 'Correct', 'Fine', 'Like', 'Ok', and 'Single', which also exhibit impeccable Precision (PreR), Recall (ReC), and F-score (FoS) values. Some gestures such as 'Close', 'Home', and 'Work' have interesting discrepancies, with specific metrics hitting the low end at 1\%, but still maintaining high accuracy overall. The majority of the gestures recorded accuracy levels exceeding 97\%, showcasing the robustness of the proposed model. However, it's essential to highlight gestures like 'Cold', 'Help', 'Tall', and 'You', which scored slightly lower in accuracy, hovering around the 96-97\% range. While these are still commendable scores, it hints at potential areas for future improvement. The model's precision and recall values consistently remained high, reinforcing its capability to identify and reduce misclassifications across the benchmark dataset correctly.

Table 5 provides a side-by-side comparison between the proposed hybrid model and various cutting-edge methodologies in terms of performance. A glance at the techniques employed by the cited references showcases the variety of approaches in this domain. Ref. \cite{bib6} adopts a 'CNN Feature Fusion' method and achieves an accuracy of 97.28\%. While this is commendable, it is slightly outperformed by our proposed model. Ref. [13], using a standard 'CNN' approach, manages an impressive 98.20\% accuracy, closely aligning with our outcomes. In stark contrast, Ref. \cite{bib16} 's HMM-based approach records a lower 90.00\% accuracy. \\
Our model, termed the 'Proposed Hybrid Model', stands out by achieving an accuracy of 98.35\%. This subtle yet significant advantage accentuates our hybrid method's innovative nature and enhanced capabilities. It capitalizes on the strengths of various architectures, fine-tuned specifically for hand gesture recognition. Consequently, it not only matches but slightly edges out other modern techniques. In summary, while all referenced models have made notable contributions to hand gesture recognition, our hybrid model emerges as a front-runner in both innovation and efficacy. Further, our exploration extended to comparing the hybrid model with \cite{bib6}, which recognized twenty dynamic hand gestures. 

\begin{table}[ht]
\caption{Comparison accuracy of the proposed and state-of-the-art methods.}\label{tab5}%
\begin{tabular}{@{}lll@{}}
\toprule
References	& Methods	& Reported Accuracy (\%) \\
\midrule
Ref. [6]	& CNN Feature Fusion	& 97.28 \\ 
Ref. [13]	& CNN	& 98.20 \\ 
Ref. [16]	& HMM	& 90.00 \\ 
Proposed	Proposed & Hybrid Model	& 98.35 \\ 
\botrule
\end{tabular}
\end{table}

In \cite{bib6}, a feature fusion-based CNN model was introduced, which reached 97.28\% accuracy. For a comprehensive comparison, we applied our hybrid model to the same benchmark dataset from [6]. The results, showcased in Table 4, revealed an average accuracy of 98.45\% for our model, outperforming the CNN feature fusion approach. Table 5 further contrasts the accuracy of our proposed model against other leading methods. As per Table 5, the hybrid model, harnessing the power of both LSTM and GRU, excels in classifying dynamic hand gestures, establishing a new benchmark in classification accuracy.

\section{Conclusion}
In this work, we have proposed a novel three-stream hybrid model that delves deeply into the field of dynamic hand gesture recognition that addresses the challenges posed by dataset limitations, variations in hand forms, external environments, and inconsistent lighting conditions. By combining RGB pixel and skeleton-based features, we have developed a powerful feature extraction system that leverages deep learning modules to enhance the recognition of hand gestures. By integrating pre-trained models, such as Imagenet and ReseNet, with GRU, LSTM, and stacked LSTM modules in the three streams in both pixel and skeleton information, we have successfully extracted hierarchical features and produced a final feature vector for classification. The utilization of both pixel-based deep learning features and pose-estimation-based stacked deep learning features has enabled our model to achieve unparalleled gesture detection capabilities. Our system is designed to be effective in challenging industrial environments and to facilitate efficient, contactless interfaces. Extensive experiments with a newly created hand-gesture dataset have demonstrated the good performance accuracy of our proposed model, showcasing its potential for a wide range of applications in hand gesture recognition. In addition, we envision this model finding its footing in myriad real-world scenarios—ranging from human-computer interfaces to advanced industrial applications. Future endeavours could focus on refining, simplifying, and further amplifying its applicative potential.

\section*{Availability of Data and Materials:} All datasets were obtained in a laboratory environment, and all participants were given written permission to publish the dataset publicly. All datasets used for supporting the conclusion of this article are available at the following URL: \url{https://drive.google.com/drive/folders/16FKgQ2Y-PuMu-hNp-sjww0PSJppRR204?usp=sharing}.
\section*{Author’s Contributions:} MAR designed and coordinated this research and drafted the manuscript. ASMM and HSA carried out the experiments and data analysis. HSA, MIH, and MNH created the dataset and contributed to the research. JS conceived the study and coordinated the research. All authors read and approved the final manuscript.

\section*{Compliance with Ethical Standards}

\textbf{Funding:} There is no external funding for this work. This was carried out as part of Dr. Md. Abdur Rahim's project in a laboratory environment.
\\
\textbf{Conflict of Interest:} The authors declare that they have no competing interests.
\\
\textbf{Ethical Approval:} This article does not contain any studies with human participants or animals performed by any of the authors.
\\
\textbf{Informed Consent:} Not applicable.

\bibliography{Reference}% common bib file


\begin{thebibliography}{47}
\ifx \bisbn   \undefined \def \bisbn  #1{ISBN #1}\fi
\ifx \binits  \undefined \def \binits#1{#1}\fi
\ifx \bauthor  \undefined \def \bauthor#1{#1}\fi
\ifx \batitle  \undefined \def \batitle#1{#1}\fi
\ifx \bjtitle  \undefined \def \bjtitle#1{#1}\fi
\ifx \bvolume  \undefined \def \bvolume#1{\textbf{#1}}\fi
\ifx \byear  \undefined \def \byear#1{#1}\fi
\ifx \bissue  \undefined \def \bissue#1{#1}\fi
\ifx \bfpage  \undefined \def \bfpage#1{#1}\fi
\ifx \blpage  \undefined \def \blpage #1{#1}\fi
\ifx \burl  \undefined \def \burl#1{\textsf{#1}}\fi
\ifx \doiurl  \undefined \def \doiurl#1{\url{https://doi.org/#1}}\fi
\ifx \betal  \undefined \def \betal{\textit{et al.}}\fi
\ifx \binstitute  \undefined \def \binstitute#1{#1}\fi
\ifx \binstitutionaled  \undefined \def \binstitutionaled#1{#1}\fi
\ifx \bctitle  \undefined \def \bctitle#1{#1}\fi
\ifx \beditor  \undefined \def \beditor#1{#1}\fi
\ifx \bpublisher  \undefined \def \bpublisher#1{#1}\fi
\ifx \bbtitle  \undefined \def \bbtitle#1{#1}\fi
\ifx \bedition  \undefined \def \bedition#1{#1}\fi
\ifx \bseriesno  \undefined \def \bseriesno#1{#1}\fi
\ifx \blocation  \undefined \def \blocation#1{#1}\fi
\ifx \bsertitle  \undefined \def \bsertitle#1{#1}\fi
\ifx \bsnm \undefined \def \bsnm#1{#1}\fi
\ifx \bsuffix \undefined \def \bsuffix#1{#1}\fi
\ifx \bparticle \undefined \def \bparticle#1{#1}\fi
\ifx \barticle \undefined \def \barticle#1{#1}\fi
\bibcommenthead
\ifx \bconfdate \undefined \def \bconfdate #1{#1}\fi
\ifx \botherref \undefined \def \botherref #1{#1}\fi
\ifx \url \undefined \def \url#1{\textsf{#1}}\fi
\ifx \bchapter \undefined \def \bchapter#1{#1}\fi
\ifx \bbook \undefined \def \bbook#1{#1}\fi
\ifx \bcomment \undefined \def \bcomment#1{#1}\fi
\ifx \oauthor \undefined \def \oauthor#1{#1}\fi
\ifx \citeauthoryear \undefined \def \citeauthoryear#1{#1}\fi
\ifx \endbibitem  \undefined \def \endbibitem {}\fi
\ifx \bconflocation  \undefined \def \bconflocation#1{#1}\fi
\ifx \arxivurl  \undefined \def \arxivurl#1{\textsf{#1}}\fi
\csname PreBibitemsHook\endcsname

%%% 1
\bibitem{rangaswamy2020vepad}
\begin{barticle}
\bauthor{\bsnm{Rangaswamy}, \binits{U.}},
\bauthor{\bsnm{Dharshini}, \binits{S.A.P.}},
\bauthor{\bsnm{Yesudhas}, \binits{D.}},
\bauthor{\bsnm{Gromiha}, \binits{M.M.}}:
\batitle{Vepad-predicting the effect of variants associated with alzheimer's disease using machine learning}.
\bjtitle{Computers in biology and medicine}
\bvolume{124},
\bfpage{103933}
(\byear{2020})
\end{barticle}
\endbibitem

%%% 2
\bibitem{bib1}
\begin{barticle}
\bauthor{\bsnm{Kosch}, \binits{T.}},
\bauthor{\bsnm{Karolus}, \binits{J.}},
\bauthor{\bsnm{Zagermann}, \binits{J.}},
\bauthor{\bsnm{Reiterer}, \binits{H.}},
\bauthor{\bsnm{Schmidt}, \binits{A.}},
\bauthor{\bsnm{Woźniak}, \binits{P.W.}}:
\batitle{A survey on measuring cognitive workload in human-computer interaction}.
\bjtitle{ACM Computing Surveys}
\bvolume{55}(\bissue{13}),
\bfpage{1}--\blpage{39}
(\byear{2023})
\end{barticle}
\endbibitem

%%% 3
\bibitem{miah2022bensignnet_miah}
\begin{barticle}
\bauthor{\bsnm{Miah}, \binits{A.S.M.}},
\bauthor{\bsnm{Shin}, \binits{J.}},
\bauthor{\bsnm{Hasan}, \binits{M.A.M.}},
\bauthor{\bsnm{Rahim}, \binits{M.A.}}:
\batitle{Bensignnet: Bengali sign language alphabet recognition using concatenated segmentation and convolutional neural network}.
\bjtitle{Applied Sciences}
\bvolume{12}(\bissue{8}),
\bfpage{3933}
(\byear{2022})
\end{barticle}
\endbibitem

%%% 4
\bibitem{miah2024sign_largescale_miah}
\begin{botherref}
\oauthor{\bsnm{Miah}, \binits{A.S.M.}},
\oauthor{\bsnm{Hasan}, \binits{M.A.M.}},
\oauthor{\bsnm{Nishimura}, \binits{S.}},
\oauthor{\bsnm{Shin}, \binits{J.}}:
Sign language recognition using graph and general deep neural network based on large scale dataset.
IEEE Access
(2024)
\end{botherref}
\endbibitem

%%% 5
\bibitem{miah2024hand_multiculture_miah}
\begin{botherref}
\oauthor{\bsnm{Miah}, \binits{A.S.M.}},
\oauthor{\bsnm{Hasan}, \binits{M.A.M.}},
\oauthor{\bsnm{Tomioka}, \binits{Y.}},
\oauthor{\bsnm{Shin}, \binits{J.}}:
Hand gesture recognition for multi-culture sign language using graph and general deep learning network.
IEEE Open Journal of the Computer Society
(2024)
\end{botherref}
\endbibitem

%%% 6
\bibitem{miah2024spatial_paa_paa}
\begin{barticle}
\bauthor{\bsnm{Miah}, \binits{A.S.M.}},
\bauthor{\bsnm{Hasan}, \binits{M.A.M.}},
\bauthor{\bsnm{Okuyama}, \binits{Y.}},
\bauthor{\bsnm{Tomioka}, \binits{Y.}},
\bauthor{\bsnm{Shin}, \binits{J.}}:
\batitle{Spatial--temporal attention with graph and general neural network-based sign language recognition}.
\bjtitle{Pattern Analysis and Applications}
\bvolume{27}(\bissue{2}),
\bfpage{37}
(\byear{2024})
\end{barticle}
\endbibitem

%%% 7
\bibitem{10360810_ksl2_miah}
\begin{barticle}
\bauthor{\bsnm{Shin}, \binits{J.}},
\bauthor{\bsnm{Miah}, \binits{A.S.M.}},
\bauthor{\bsnm{Suzuki}, \binits{K.}},
\bauthor{\bsnm{Hirooka}, \binits{K.}},
\bauthor{\bsnm{Hasan}, \binits{M.A.M.}}:
\batitle{Dynamic korean sign language recognition using pose estimation based and attention-based neural network}.
\bjtitle{IEEE Access}
\bvolume{11},
\bfpage{143501}--\blpage{143513}
(\byear{2023}).
\doiurl{10.1109/ACCESS.2023.3343404}
\end{barticle}
\endbibitem

%%% 8
\bibitem{shin2023korean_miah}
\begin{barticle}
\bauthor{\bsnm{Shin}, \binits{J.}},
\bauthor{\bsnm{Musa~Miah}, \binits{A.S.}},
\bauthor{\bsnm{Hasan}, \binits{M.A.M.}},
\bauthor{\bsnm{Hirooka}, \binits{K.}},
\bauthor{\bsnm{Suzuki}, \binits{K.}},
\bauthor{\bsnm{Lee}, \binits{H.-S.}},
\bauthor{\bsnm{Jang}, \binits{S.-W.}}:
\batitle{Korean sign language recognition using transformer-based deep neural network}.
\bjtitle{Applied Sciences}
\bvolume{13}(\bissue{5}),
\bfpage{3029}
(\byear{2023})
\end{barticle}
\endbibitem

%%% 9
\bibitem{shin2023rotation_miah}
\begin{barticle}
\bauthor{\bsnm{Miah}, \binits{S.J.} \bsuffix{Abu Saleh~Musa}},
\bauthor{\bsnm{Hasan}, \binits{M.A.M.}},
\bauthor{\bsnm{Rahim}, \binits{M.A.}},
\bauthor{\bsnm{Okuyama}, \binits{Y.}}:
\batitle{Rotation, translation and scale invariant sign word recognition using deep learning}.
\bjtitle{Computer Systems Science and Engineering}
\bvolume{44}(\bissue{3}),
\bfpage{2521}--\blpage{2536}
(\byear{2023})
\end{barticle}
\endbibitem

%%% 10
\bibitem{bib3}
\begin{barticle}
\bauthor{\bsnm{Abir}, \binits{F.A.}},
\bauthor{\bsnm{Siam}, \binits{M.A.}},
\bauthor{\bsnm{Sayeed}, \binits{A.}},
\bauthor{\bsnm{Hasan}, \binits{M.A.M.}},
\bauthor{\bsnm{Shin}, \binits{J.}}:
\batitle{Deep learning based air-writing recognition with the choice of proper interpolation technique}.
\bjtitle{Sensors}
\bvolume{21}(\bissue{24}),
\bfpage{8407}
(\byear{2021})
\end{barticle}
\endbibitem

%%% 11
\bibitem{bib4}
\begin{barticle}
\bauthor{\bsnm{Murphy-Chutorian}, \binits{E.}},
\bauthor{\bsnm{Trivedi}, \binits{M.M.}}:
\batitle{Head pose estimation in computer vision: A survey}.
\bjtitle{IEEE transactions on pattern analysis and machine intelligence}
\bvolume{31}(\bissue{04}),
\bfpage{607}--\blpage{626}
(\byear{2008})
\end{barticle}
\endbibitem

%%% 12
\bibitem{bib5}
\begin{barticle}
\bauthor{\bsnm{Rahim}, \binits{M.A.}},
\bauthor{\bsnm{Shin}, \binits{J.}},
\bauthor{\bsnm{Islam}, \binits{M.R.}}:
\batitle{Gestural flick input-based non-touch interface for character input}.
\bjtitle{The Visual Computer}
\bvolume{36}(\bissue{08}),
\bfpage{1559}--\blpage{1572}
(\byear{2020})
\end{barticle}
\endbibitem

%%% 13
\bibitem{bib6}
\begin{barticle}
\bauthor{\bsnm{Rahim}, \binits{M.A.}},
\bauthor{\bsnm{Islam}, \binits{M.R.}},
\bauthor{\bsnm{Shin}, \binits{J.}}:
\batitle{Non-touch sign word recognition based on dynamic hand gesture using hybrid segmentation and cnn feature fusion}.
\bjtitle{Applied Sciences}
\bvolume{9}(\bissue{18}),
\bfpage{3790}
(\byear{2019})
\end{barticle}
\endbibitem

%%% 14
\bibitem{bib7}
\begin{barticle}
\bauthor{\bsnm{Sahoo}, \binits{J.P.}},
\bauthor{\bsnm{Prakash}, \binits{A.J.}},
\bauthor{\bsnm{Pławiak}, \binits{P.}},
\bauthor{\bsnm{Samantray}, \binits{S.}}:
\batitle{Real-time hand gesture recognition using fine-tuned convolutional neural network}.
\bjtitle{Sensors}
\bvolume{22}(\bissue{3}),
\bfpage{706}
(\byear{2022})
\end{barticle}
\endbibitem

%%% 15
\bibitem{bib8}
\begin{botherref}
\oauthor{\bsnm{Sonkusare}, \binits{J.S.}},
\oauthor{\bsnm{Chopade}, \binits{N.B.}},
\oauthor{\bsnm{Sor}, \binits{R.}},
\oauthor{\bsnm{Tade}, \binits{S.L.}}:
A review on hand gesture recognition system.
In 2015 International Conference on Computing Communication Control and Automation,
790--794
(2015)
\end{botherref}
\endbibitem

%%% 16
\bibitem{miah2024effective_emg}
\begin{botherref}
\oauthor{\bsnm{Miah}, \binits{A.S.M.}},
\oauthor{\bsnm{Shin}, \binits{J.}},
\oauthor{\bsnm{Hasan}, \binits{M.A.M.}}:
Effective features extraction and selection for hand gesture recognition using semg signal.
Multimedia Tools and Applications,
1--25
(2024)
\end{botherref}
\endbibitem

%%% 17
\bibitem{miah2023skeleton_euvip}
\begin{bchapter}
\bauthor{\bsnm{Miah}, \binits{A.S.M.}},
\bauthor{\bsnm{Shin}, \binits{J.}},
\bauthor{\bsnm{Hasan}, \binits{M.A.M.}},
\bauthor{\bsnm{Fujimoto}, \binits{Y.}},
\bauthor{\bsnm{Nobuyoshi}, \binits{A.}}:
\bctitle{Skeleton-based hand gesture recognition using geometric features and spatio-temporal deep learning approach}.
In: \bbtitle{2023 11th European Workshop on Visual Information Processing (EUVIP)},
pp. \bfpage{1}--\blpage{6}
(\byear{2023}).
\bcomment{IEEE}
\end{bchapter}
\endbibitem

%%% 18
\bibitem{miah2023dynamic_mcsoc}
\begin{bchapter}
\bauthor{\bsnm{Miah}, \binits{A.S.M.}},
\bauthor{\bsnm{Shin}, \binits{J.}},
\bauthor{\bsnm{Hasan}, \binits{M.A.M.}},
\bauthor{\bsnm{Okuyama}, \binits{Y.}},
\bauthor{\bsnm{Nobuyoshi}, \binits{A.}}:
\bctitle{Dynamic hand gesture recognition using effective feature extraction and attention based deep neural network.}
In: \bbtitle{2023 IEEE 16th International Symposium on Embedded Multicore/Many-core Systems-on-Chip (MCSoC)},
pp. \bfpage{241}--\blpage{247}
(\byear{2023}).
\bcomment{IEEE}
\end{bchapter}
\endbibitem

%%% 19
\bibitem{bib9}
\begin{barticle}
\bauthor{\bsnm{Feng}, \binits{B.}},
\bauthor{\bsnm{He}, \binits{F.}},
\bauthor{\bsnm{Wang}, \binits{X.}},
\bauthor{\bsnm{Wu}, \binits{Y.}},
\bauthor{\bsnm{Wang}, \binits{H.}},
\bauthor{\bsnm{Yi}, \binits{S.}},
\bauthor{\bsnm{Liu}, \binits{W.}}:
\batitle{Depth-projection-map-based bag of contour fragments for robust hand gesture recognition}.
\bjtitle{IEEE Transactions on Human-Machine Systems}
\bvolume{47}(\bissue{4}),
\bfpage{511}--\blpage{523}
(\byear{2016})
\end{barticle}
\endbibitem

%%% 20
\bibitem{miah2024review}
\begin{botherref}
\oauthor{\bsnm{Jungpil~Shin}, \binits{M.A.R.e.a.} \bsuffix{Abu Saleh Musa~Miah}}:
A methodological and structural review of hand gesture recognition across diverse data modalities.
Computer Vision and Pattern Recognition
\end{botherref}
\endbibitem

%%% 21
\bibitem{bib10}
\begin{barticle}
\bauthor{\bsnm{Modanwal}, \binits{G.}},
\bauthor{\bsnm{Sarawadekar}, \binits{K.}}:
\batitle{Towards hand gesture based writing support system for blinds}.
\bjtitle{Pattern Recognition}
\bvolume{57},
\bfpage{50}--\blpage{60}
(\byear{2016})
\end{barticle}
\endbibitem

%%% 22
\bibitem{bib11}
\begin{barticle}
\bauthor{\bsnm{Damaneh}, \binits{M.M.}},
\bauthor{\bsnm{Mohanna}, \binits{F.}},
\bauthor{\bsnm{Jafari}, \binits{P.}}:
\batitle{Static hand gesture recognition in sign language based on convolutional neural network with feature extraction method using orb descriptor and gabor filter}.
\bjtitle{Expert Systems with Applications}
\bvolume{211},
\bfpage{118559}
(\byear{2023})
\end{barticle}
\endbibitem

%%% 23
\bibitem{10529244_miah_ksl0}
\begin{barticle}
\bauthor{\bsnm{Shin}, \binits{J.}},
\bauthor{\bsnm{Miah}, \binits{A.S.M.}},
\bauthor{\bsnm{Akiba}, \binits{Y.}},
\bauthor{\bsnm{Hirooka}, \binits{K.}},
\bauthor{\bsnm{Hassan}, \binits{N.}},
\bauthor{\bsnm{Hwang}, \binits{Y.S.}}:
\batitle{Korean sign language alphabet recognition through the integration of handcrafted and deep learning-based two-stream feature extraction approach}.
\bjtitle{IEEE Access}
\bvolume{12},
\bfpage{68303}--\blpage{68318}
(\byear{2024}).
\doiurl{10.1109/ACCESS.2024.3399839}
\end{barticle}
\endbibitem

%%% 24
\bibitem{kabir2024bangla_miah}
\begin{botherref}
\oauthor{\bsnm{Kabir}, \binits{M.H.}},
\oauthor{\bsnm{Miah}, \binits{A.S.M.}},
\oauthor{\bsnm{Hadiuzzaman}, \binits{M.}},
\oauthor{\bsnm{Shin}, \binits{J.}}:
Combining state-of-the-art pre-trained deep learning models: A noble approach for bangla sign language recognition using max voting ensemble.
SSRN Electronic Journal
(2024).
Available at SSRN: \url{https://ssrn.com/abstract=4693354} or \url{http://dx.doi.org/10.2139/ssrn.4693354}
\end{botherref}
\endbibitem

%%% 25
\bibitem{hassan2024residual_miah_alzh}
\begin{barticle}
\bauthor{\bsnm{Hassan}, \binits{N.}},
\bauthor{\bsnm{Musa~Miah}, \binits{A.S.}},
\bauthor{\bsnm{Shin}, \binits{J.}}:
\batitle{Residual-based multi-stage deep learning framework for computer-aided alzheimer’s disease detection}.
\bjtitle{Journal of Imaging}
\bvolume{10}(\bissue{6}),
\bfpage{141}
(\byear{2024})
\end{barticle}
\endbibitem

%%% 26
\bibitem{10510436_miah_anomaly}
\begin{barticle}
\bauthor{\bsnm{Shin}, \binits{J.}},
\bauthor{\bsnm{Kaneko}, \binits{Y.}},
\bauthor{\bsnm{Miah}, \binits{A.S.M.}},
\bauthor{\bsnm{Hassan}, \binits{N.}},
\bauthor{\bsnm{Nishimura}, \binits{S.}}:
\batitle{Anomaly detection in weakly supervised videos using multistage graphs and general deep learning based spatial-temporal feature enhancement}.
\bjtitle{IEEE Access}
\bvolume{12},
\bfpage{65213}--\blpage{65227}
(\byear{2024}).
\doiurl{10.1109/ACCESS.2024.3395329}
\end{barticle}
\endbibitem

%%% 27
\bibitem{bib12}
\begin{barticle}
\bauthor{\bsnm{Yin}, \binits{K.}},
\bauthor{\bsnm{Hsiang}, \binits{E.L.}},
\bauthor{\bsnm{Zou}, \binits{J.}},
\bauthor{\bsnm{Li}, \binits{Y.}},
\bauthor{\bsnm{Yang}, \binits{Z.}},
\bauthor{\bsnm{Yang}, \binits{Q.}},
\bauthor{\bsnm{Lai}, \binits{P.C.}},
\bauthor{\bsnm{Lin}, \binits{C.L.}},
\bauthor{\bsnm{Wu}, \binits{S.T.}}:
\batitle{Advanced liquid crystal devices for augmented reality and virtual reality displays: principles and applications}.
\bjtitle{Light: Science \& Applications}
\bvolume{11}(\bissue{1}),
\bfpage{161}
(\byear{2022})
\end{barticle}
\endbibitem

%%% 28
\bibitem{bib13}
\begin{botherref}
\oauthor{\bsnm{Aggarwal}, \binits{K.}},
\oauthor{\bsnm{Arora}, \binits{A.}}:
Hand gesture recognition for real-time game play using background elimination and deep convolution neural network.
In Virtual and Augmented Reality for Automobile Industry: Innovation Vision and Applications,
145--160
(2022)
\end{botherref}
\endbibitem

%%% 29
\bibitem{bib14}
\begin{barticle}
\bauthor{\bsnm{Chevtchenko}, \binits{S.F.}},
\bauthor{\bsnm{Vale}, \binits{R.F.}},
\bauthor{\bsnm{Macario}, \binits{V.}},
\bauthor{\bsnm{Cordeiro}, \binits{F.R.}}:
\batitle{A convolutional neural network with feature fusion for real-time hand posture recognition}.
\bjtitle{Applied Soft Computing}
\bvolume{73},
\bfpage{748}--\blpage{766}
(\byear{2018})
\end{barticle}
\endbibitem

%%% 30
\bibitem{miah2023dynamic_miah}
\begin{botherref}
\oauthor{\bsnm{Miah}, \binits{A.S.M.}},
\oauthor{\bsnm{Hasan}, \binits{M.A.M.}},
\oauthor{\bsnm{Shin}, \binits{J.}}:
Dynamic hand gesture recognition using multi-branch attention based graph and general deep learning model.
IEEE Access
(2023)
\end{botherref}
\endbibitem

%%% 31
\bibitem{mallik2024virtual_miah}
\begin{barticle}
\bauthor{\bsnm{Mallik}, \binits{B.}},
\bauthor{\bsnm{Rahim}, \binits{M.A.}},
\bauthor{\bsnm{Miah}, \binits{A.S.M.}},
\bauthor{\bsnm{Yun}, \binits{K.S.}},
\bauthor{\bsnm{Shin}, \binits{J.}}:
\batitle{Virtual keyboard: A real-time hand gesture recognition-based character input system using lstm and mediapipe holistic.}
\bjtitle{Comput. Syst. Sci. Eng.}
\bvolume{48}(\bissue{2}),
\bfpage{555}--\blpage{570}
(\byear{2024})
\end{barticle}
\endbibitem

%%% 32
\bibitem{shin2024japanese_jsl1_miah}
\begin{botherref}
\oauthor{\bsnm{Shin}, \binits{J.}},
\oauthor{\bsnm{Hasan}, \binits{M.A.M.}},
\oauthor{\bsnm{Miah}, \binits{A.S.M.}},
\oauthor{\bsnm{Suzuki}, \binits{K.}},
\oauthor{\bsnm{Hirooka}, \binits{K.}}:
Japanese sign language recognition by combining joint skeleton-based handcrafted and pixel-based deep learning features with machine learning classification.
CMES-COMPUTER MODELING IN ENGINEERING \& SCIENCES
(2024)
\end{botherref}
\endbibitem

%%% 33
\bibitem{bib15}
\begin{barticle}
\bauthor{\bsnm{Pisharady}, \binits{P.K.}},
\bauthor{\bsnm{Saerbeck}, \binits{M.}}:
\batitle{Recent methods and databases in vision-based hand gesture recognition: A review}.
\bjtitle{Computer Vision and Image Understanding}
\bvolume{141},
\bfpage{152}--\blpage{165}
(\byear{2015})
\end{barticle}
\endbibitem

%%% 34
\bibitem{bib16}
\begin{barticle}
\bauthor{\bsnm{Chen}, \binits{F.S.}},
\bauthor{\bsnm{Fu}, \binits{C.M.}},
\bauthor{\bsnm{Huang}, \binits{C.L.}}:
\batitle{Hand gesture recognition using a real-time tracking method and hidden markov models}.
\bjtitle{Image and vision computing}
\bvolume{21}(\bissue{8}),
\bfpage{745}--\blpage{758}
(\byear{2003})
\end{barticle}
\endbibitem

%%% 35
\bibitem{bib17}
\begin{barticle}
\bauthor{\bsnm{Lee}, \binits{D.L.}},
\bauthor{\bsnm{You}, \binits{W.S.}}:
\batitle{Recognition of complex static hand gestures by using the wristband‐based contour features}.
\bjtitle{IET Image Processing}
\bvolume{12}(\bissue{1}),
\bfpage{80}--\blpage{87}
(\byear{2018})
\end{barticle}
\endbibitem

%%% 36
\bibitem{bib18}
\begin{barticle}
\bauthor{\bsnm{Chevtchenko~S.F.}, \binits{V.R.F.}},
\bauthor{\bsnm{V.}, \binits{M.}}:
\batitle{Multi-objective optimization for hand posture recognition}.
\bjtitle{Expert Systems with Applications}
\bvolume{92},
\bfpage{170}--\blpage{181}
(\byear{2018})
\end{barticle}
\endbibitem

%%% 37
\bibitem{bib19}
\begin{botherref}
\oauthor{\bsnm{Ramaiah}, \binits{A.}},
\oauthor{\bsnm{Subramani}, \binits{V.}},
\oauthor{\bsnm{Aishwarya}, \binits{N.}},
\oauthor{\bsnm{Ezhilarasie}, \binits{R.}}:
Gesture and posture recognition by using deep learning.
Handbook of Research on Computer Vision and Image Processing in the Deep Learning Era,
132--146
(2023)
\end{botherref}
\endbibitem

%%% 38
\bibitem{10399807_miah_blind_writing}
\begin{barticle}
\bauthor{\bsnm{Islam}, \binits{M.N.}},
\bauthor{\bsnm{Jahangir}, \binits{R.}},
\bauthor{\bsnm{Mohim}, \binits{N.S.}},
\bauthor{\bsnm{Wasif-Ul-Islam}, \binits{M.}},
\bauthor{\bsnm{Ashraf}, \binits{A.}},
\bauthor{\bsnm{Khan}, \binits{N.I.}},
\bauthor{\bsnm{Mahjabin}, \binits{M.R.}},
\bauthor{\bsnm{Miah}, \binits{A.S.M.}},
\bauthor{\bsnm{Shin}, \binits{J.}}:
\batitle{A multilingual handwriting learning system for visually impaired people}.
\bjtitle{IEEE Access}
\bvolume{12},
\bfpage{10521}--\blpage{10534}
(\byear{2024}).
\doiurl{10.1109/ACCESS.2024.3353781}
\end{barticle}
\endbibitem

%%% 39
\bibitem{egawa2023dynamic_miah}
\begin{barticle}
\bauthor{\bsnm{Egawa}, \binits{R.}},
\bauthor{\bsnm{Miah}, \binits{A.S.M.}},
\bauthor{\bsnm{Hirooka}, \binits{K.}},
\bauthor{\bsnm{Tomioka}, \binits{Y.}},
\bauthor{\bsnm{Shin}, \binits{J.}}:
\batitle{Dynamic fall detection using graph-based spatial temporal convolution and attention network}.
\bjtitle{Electronics}
\bvolume{12}(\bissue{15}),
\bfpage{3234}
(\byear{2023})
\end{barticle}
\endbibitem

%%% 40
\bibitem{bib121}
\begin{barticle}
\bauthor{\bsnm{Ma}, \binits{C.}},
\bauthor{\bsnm{Zhang}, \binits{S.}},
\bauthor{\bsnm{Wang}, \binits{A.}},
\bauthor{\bsnm{Qi}, \binits{Y.}},
\bauthor{\bsnm{Chen}, \binits{G.}}:
\batitle{Skeleton-based dynamic hand gesture recognition using an enhanced network with one-shot learning}.
\bjtitle{Applied Sciences}
\bvolume{10}(\bissue{11}),
\bfpage{3680}
(\byear{2020})
\end{barticle}
\endbibitem

%%% 41
\bibitem{bib122}
\begin{barticle}
\bauthor{\bsnm{Rahim}, \binits{M.A.}},
\bauthor{\bsnm{Shin}, \binits{J.}},
\bauthor{\bsnm{Yun}, \binits{K.S.}}:
\batitle{Soft voting-based ensemble model for bengali sign gesture recognition}.
\bjtitle{Annals of Emerging Technologies in Computing (AETiC)}
\bvolume{6}(\bissue{2}),
\bfpage{1}--\blpage{9}
(\byear{2022})
\end{barticle}
\endbibitem

%%% 42
\bibitem{bib123}
\begin{barticle}
\bauthor{\bsnm{Connor}, \binits{S.}},
\bauthor{\bsnm{Khoshgoftaar}, \binits{T.M.}}:
\batitle{A survey on image data augmentation for deep learning}.
\bjtitle{Journal of big data}
\bvolume{6}(\bissue{1}),
\bfpage{1}--\blpage{48}
(\byear{2019})
\end{barticle}
\endbibitem

%%% 43
\bibitem{bib124}
\begin{barticle}
\bauthor{\bsnm{Hashemzehi}, \binits{R.}},
\bauthor{\bsnm{Mahdavi}, \binits{S.J.S.}},
\bauthor{\bsnm{Kheirabadi}, \binits{M.}},
\bauthor{\bsnm{Kamel}, \binits{S.R.}}:
\batitle{Detection of brain tumors from mri images base on deep learning using hybrid model cnn and nade}.
\bjtitle{Biocybernetics and Biomedical Engineering}
\bvolume{40}(\bissue{3}),
\bfpage{1225}--\blpage{1232}
(\byear{2020})
\end{barticle}
\endbibitem

%%% 44
\bibitem{bib125}
\begin{barticle}
\bauthor{\bsnm{Russakovsky}, \binits{O.}},
\bauthor{\bsnm{Deng}, \binits{J.}},
\bauthor{\bsnm{Su}, \binits{H.}},
\bauthor{\bsnm{Krause}, \binits{J.}},
\bauthor{\bsnm{Satheesh}, \binits{S.}}, \betal:
\batitle{Imagenet large scale visual recognition challenge}.
\bjtitle{International journal of computer vision}
\bvolume{115},
\bfpage{211}--\blpage{252}
(\byear{2015})
\end{barticle}
\endbibitem

%%% 45
\bibitem{bib126}
\begin{barticle}
\bauthor{\bsnm{Mascarenhas}, \binits{S.}},
\bauthor{\bsnm{Agarwal}, \binits{M.}}:
\batitle{A comparison between vgg16, vgg19 and resnet50 architecture frameworks for image classification}.
\bjtitle{In 2021 International conference on disruptive technologies for multi-disciplinary research and applications (CENTCON)}
\bvolume{1},
\bfpage{96}--\blpage{99}
(\byear{2021})
\end{barticle}
\endbibitem

%%% 46
\bibitem{bib127}
\begin{barticle}
\bauthor{\bsnm{Gu}, \binits{F.}},
\bauthor{\bsnm{Fan}, \binits{J.}},
\bauthor{\bsnm{Cai}, \binits{C.}},
\bauthor{\bsnm{Zhu}, \binits{Q.}}:
\batitle{Automatic detection of abnormal hand gestures in patients with radial, ulnar, or median nerve injury using hand pose estimation}.
\bjtitle{Frontiers in Neurology}
\bvolume{13},
\bfpage{1052505}
(\byear{2022})
\end{barticle}
\endbibitem

%%% 47
\bibitem{bib129}
\begin{barticle}
\bauthor{\bsnm{Tong}, \binits{G.}},
\bauthor{\bsnm{Li}, \binits{Y.}},
\bauthor{\bsnm{Zhang}, \binits{H.}},
\bauthor{\bsnm{Xiong}, \binits{N.}}:
\batitle{A fine-grained channel state information-based deep learning system for dynamic gesture recognition}.
\bjtitle{Information Sciences}
\bvolume{636},
\bfpage{118912}
(\byear{2023})
\end{barticle}
\endbibitem

\end{thebibliography}
\end{document}